\documentclass{Interspeech}

\interspeechcameraready

\title{LiSTEN: Learning Soft Token Embeddings for Neural Audio LLMs}

\author[affiliation={1,3}]{Pooneh}{Mousavi*}
\author[affiliation={2,3}]{Shubham}{Gupta*}
\author[affiliation={2,3}]{Cem}{Subakan}
\author[affiliation={1,3}]{Mirco}{Ravanelli}

\affiliation{}{Concordia University}{Canada}
\affiliation{}{Université Laval}{Canada}
\affiliation{}{Mila - Quebec AI Institute}{Canada}
\email{pooneh.mousavi@mail.concordia.ca, shgup1@ulaval.ca}

\usepackage{comment}
\usepackage{amsmath}
\usepackage{stfloats}

\begin{document}

\maketitle
\begingroup
\renewcommand\thefootnote{*}
\footnotetext{Equal contribution, Accepted at Interspeech 2025}
\endgroup
\begin{abstract}

Foundation models based on large language models (LLMs) have shown great success in handling various tasks and modalities. However, adapting these models for general-purpose audio-language tasks is challenging due to differences in acoustic environments and task variations.  In this work, we introduce \textbf{LiSTEN} (\textbf{L}earning \textbf{S}oft \textbf{T}oken \textbf{E}mbeddings for \textbf{N}eural Audio LLMs), a framework for adapting LLMs to speech and audio tasks. LiSTEN uses a dynamic prompt selection strategy with learnable key-value pairs, allowing the model to balance general and task-specific knowledge while avoiding overfitting in a multitask setting.  Our approach reduces dependence on large-scale ASR or captioning datasets, achieves competitive performance with fewer trainable parameters, and simplifies training by using a single-stage process. Additionally, LiSTEN enhances interpretability by analyzing the diversity and overlap of selected prompts across different tasks.

\end{abstract}

\section{Introduction}
Foundation models built on large language models (LLMs) have demonstrated strong capabilities in handling diverse tasks and modalities \cite{deshmukh2023pengi,tang2023salmonn,chu2024qwen2,zhan2024anygpt,xie2024mini,gong2023listen,ma2024languagemodellistenspeaking, kong2024audioflamingonovelaudio,lu2024destaenhancingspeechlanguage,deshmukh2024pengiaudiolanguagemodel,rubenstein2023audiopalm}. In recent years, text-based LLMs have achieved remarkable performance, serving as the backbone for multimodal systems. A key factor behind their success is \textit{instruction tuning} \cite{peng2023instruction,zhang2023instruction, taori2023alpaca}, where models are trained on instruction-response pairs to enhance their ability to follow diverse prompts.

Expanding LLMs to process auditory inputs introduces unique challenges, particularly in adapting to different acoustic environments and task variations. \textit{Audio-language models (ALMs)} address this by integrating pre-trained audio encoders with adapter modules that align different input modalities. A common strategy for adapting LLMs to new modalities is \textit{parameter-efficient fine-tuning (PEFT)}. Techniques such as Low Rank Adaptation (LoRA) \cite{hu2021lora,dettmers2024qlora} enable parameter-efficient adaptation by updating only a small subset of parameters while keeping most of the model unchanged, reducing computational costs \cite{tang2023salmonn,chu2024qwen2,hu2024wavllm, gong2023listen, lu2024destaenhancingspeechlanguage}. Another approach, \textit{prompt tuning} \cite{lester2021power,li2021prefix, gupta2025stopmodelingasynchronoustime}, optimizes a small set of learnable prompts to guide model behavior without modifying its core structure \cite{chang2023speechprompt, hu2024chain, wu2023speechgen}. 

Despite these advancements, training \textit{Audio LLMs (ALLMs)} remains challenging, requiring large and diverse datasets as well as significant engineering efforts. These models often rely on cross-modal learning and instruction tuning to improve their ability to integrate multimodal data effectively. Many models, such as \textit{SALMONN} \cite{tang2023salmonn}, \textit{WAVLLM} \cite{hu2024wavllm}, and \textit{Qwen} \cite{chu2024qwen2}, employ a multi-stage training approach to enhance the fusion of audio and text data. The first stage typically focuses on aligning audio and text representations using large-scale datasets for automatic speech recognition (ASR) or audio captioning (ACAP), followed by fine-tuning for specific tasks. However, heavy reliance on such datasets can lead to \textit{task overfitting}, where models perform well on trained tasks but struggle to generalize to new ones. This issue is particularly problematic in \textit{few-shot learning} (tasks with limited data) and \textit{zero-shot learning} (tasks without training data).  Additionally, LoRA-based fine-tuning can result in the loss of text-based commonsense knowledge, which is crucial for spoken question-answering (SQA) tasks. SpeechPrompt\cite{chang2022speechpromptexplorationprompttuning, chang2023speechprompt} attempts to address this issue by training task-specific prompts. However, this approach compromises the shared knowledge across multiple tasks and eliminates the zero-shot generalization benefits provided by multitask LoRA fine-tuning methods such as SALMONN and Qwen. Moreover, maintaining separate prompts for each task reduces the model’s scalability to new tasks, as each new task requires a separately trained and stored prompt. 

To address these challenges, we propose \textbf{LiSTEN}: \textbf{L}earning \textbf{S}oft \textbf{T}oken \textbf{E}mbeddings for \textbf{N}eural Audio LLMs a novel framework for efficiently adapting LLMs to process auditory inputs. LiSTEN integrates a pre-trained text-based LLM with speech and audio encoders and employs a \textit{dynamic prompt selection (DPS)} strategy. Unlike traditional fine-tuning, LiSTEN selects \textit{learnable prompts} dynamically based on speech input and task instructions, eliminating the need for manual selection or fully shared parameters across multiple tasks, which can lead to overfitting and forgetting. By continuously optimizing prompts, LiSTEN balances task-invariant and task-specific knowledge, enabling adaptation to new tasks and domains while maintaining prior knowledge.

Our key contributions are as follows:

\begin{itemize}
    \item We introduce LiSTEN, a dynamic prompt-based fine-tuning framework for efficiently adapting LLMs to speech and audio tasks.
    \item LiSTEN eliminates the need for large-scale ASR or captioning datasets, enabling adaptation with significantly less training data. We demonstrate that our approach achieves competitive performance with a single-stage training process and fewer trainable parameters.
    \item Our dynamic prompt selection inherently enhances explainability by selecting similar prompts for related tasks while choosing diverse prompts when task objectives or input data distributions differ.
\end{itemize}

Our results highlight the potential of \textit{dynamic prompt learning} for scalable, adaptive audio LLM, enabling AI systems to efficiently process and respond to a diverse range of  queries.

\section{Model Design}
\begin{figure*}[t]
    \centering
    \scalebox{0.9}{
    \includegraphics[width=0.8\textwidth]{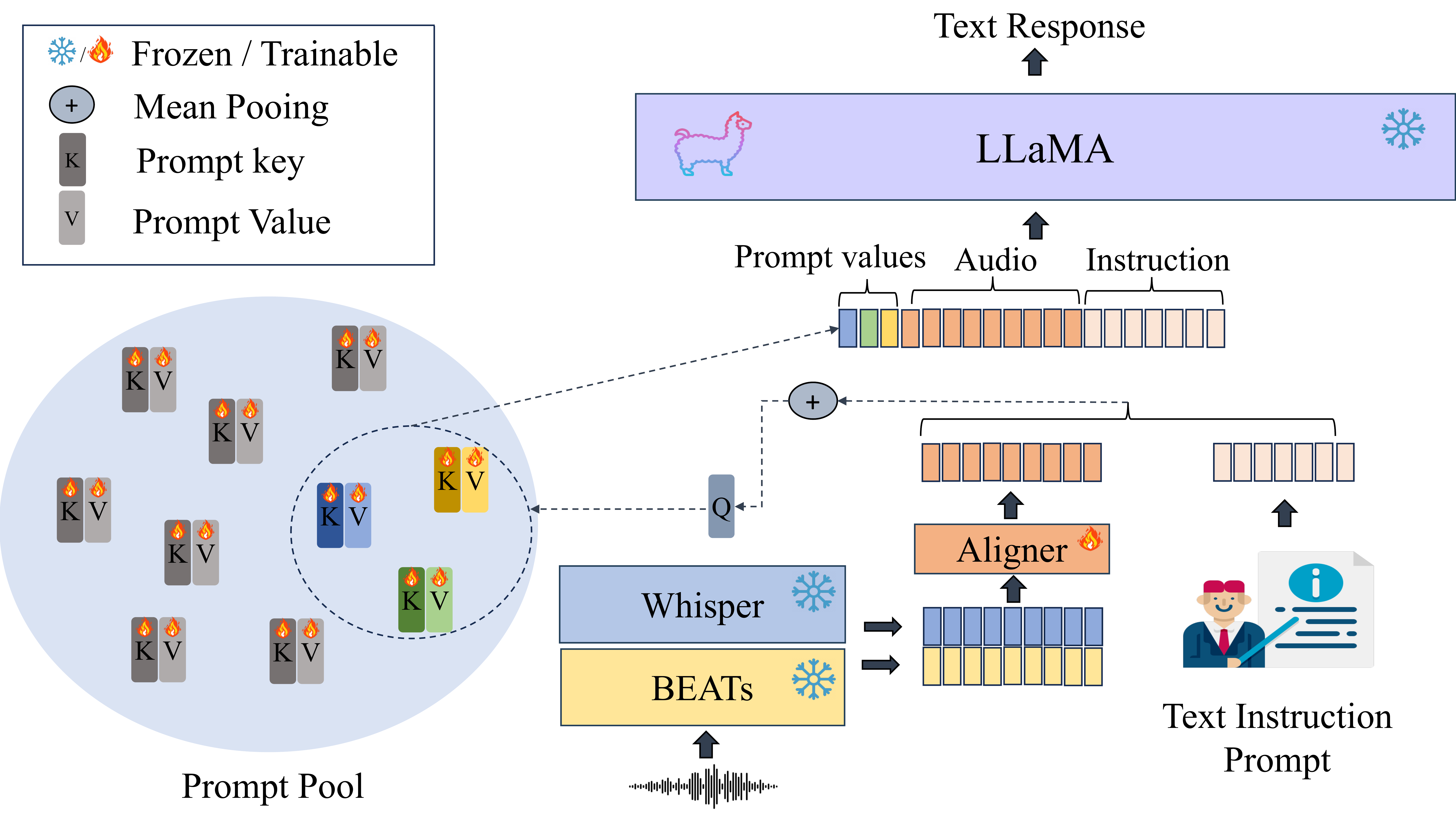}
    }
    \caption{LiSTEN pipeline with Dynamic Prompt Selection (DPS). The prompt pool consists of key-value pairs. Audio is encoded using Whisper and BEATs, whose embeddings are concatenated and processed through a Q-Former. The task is represented as text and encoded with the backbone LLM tokenizer. The speech and text tokens are mean-pooled to obtain a query, which selects $k$ values from the prompt pool. These selected values serve as the soft prompt for the instance and are prepended to the input before being passed to the LLM.}

    \label{fig:listen_pipeline}
\end{figure*}

We introduce a new fine-tuning technique for training Audio Large Language Models (ALLM). Inspired by L2P \cite{wang2022learning,wang2022dualprompt}, we implement a prompt selection mechanism where prompts are stored as learnable key-value pairs. Our proposed architecture is illustrated in Figure \ref{fig:listen_pipeline}. 

Our setup follows a similar structure to SALMONN\footnote{\url{https://huggingface.co/tsinghua-ee/SALMONN}}, but differs in its fine-tuning approach. Like SALMONN, we integrate two audio encoders: OpenAI’s Whisper \cite{radford2023robust} for speech processing and BEATs \cite{chen2022beats} for non-speech sounds. Whisper is trained on large-scale weakly supervised data for speech recognition and translation, while BEATs captures high-level non-speech audio features through self-supervised learning. Since both encoders output at a 50Hz frame rate, we concatenate their features to create a unified audio representation.

To handle variable-length audio inputs, we adapt the Q-Former structure \cite{li2023blip}, originally designed for images, to process audio at the window level instead of the full sequence. The encoder output is divided into fixed-size windows, where Q-Former processes each window separately and converts them into textual tokens for the language model. This approach improves efficiency while ensuring adaptability to different audio lengths.

For the language model, we use a pre-trained LLaMA LLM \cite{touvron2023llama}. Instead of fine-tuning with LoRA, a common parameter-efficient adaptation method, we introduce a prompt pooling approach. After extracting features from Q-Former and combining them with text instructions, we compute an average over the entire sequence to generate a query vector. This vector captures information from both speech and text sequences, representing the task and data distribution. We then compare this query vector with prompt keys stored in a prompt pool, selecting the top-k matching prompts based on various selection criteria. The corresponding prompt values are retrieved and prepended to the extracted features before being passed to the LLM. The only trainable parameters in our method are the Q-Former, which aligns audio features to LLM dimensions, and the prompt keys and values which allow task-adaptive prompting.
We explore different prompt selection strategies in the next section.

\subsection{Prompt Selection Strategies}
\label{prompt_selection_stratgies}

For prompt selection, we first condense the input into a dense representation that serves as a query for the prompt pool:   $q = \frac{1}{T} \sum_{t=1}^T X_t$, where $X \in \mathbb{R}^{T \times d}$ represents the concatenation of speech and text tokens, with \(T\) being the total number of tokens after concatenation (each of dimension \(d\)). This vector \(q\) captures task-relevant information and is used to retrieve prompts from the prompt pool. The prompt pool is defined as  $\{(k_i, v_i)\}_{i=1}^P$, with each key \(k_i \in \mathbb{R}^d\) and corresponding value \(v_i \in \mathbb{R}^{d'}\) being learnable parameters. Our goal is to select \(k\) prompt values based on \(q\) using one of the following strategies.
\\
\\
\noindent \textbf{Similarity-Based Selection: } To determine the most relevant prompts, we first compute the cosine similarity between the input representation $q$ and each prompt key $k_i$ as  
$\text{sim}(q, k_i) = \frac{q \cdot k_i}{\lVert q \rVert \, \lVert k_i \rVert}$.  
We apply a softmax function:  
$\alpha_i = \frac{\exp(\text{sim}(q, k_i))}{\sum_{j=1}^{P} \exp(\text{sim}(q, k_j))}$ to normalize these scores between $0$ and $1$. Finally, we select the top-$k$ keys with the highest scores, and their corresponding prompt values are used.
\\
\\
\noindent \textbf{Attention-Based Selection:} Treat $q$ as a query to compute attention scores for each prompt key:  
$\alpha_i = \frac{\exp(q^\top k_i)}{\sum_{j=1}^{P} \exp(q^\top k_j)}$.  
The final prompt is obtained by selecting the values corresponding to the top-$k$ keys with the highest attention scores. These values are then scaled by their respective attention weights, yielding the final prompt as:  
$v_i' = \alpha_i v_i$ for $i \in \mathcal{K}$,  
where $\mathcal{K}$ is the set of indices corresponding to the top-$k$ keys. This approach enables a soft, dynamic weighting of prompt values rather than a hard selection, allowing multiple prompts to contribute in proportion to their relevance.
\\
\\
\noindent \textbf{Residual-Based Selection:} Inspired by the residual vector quantization (RVQ) approach introduced in \cite{9625818}, this method follows an iterative process. First, we select the key closest to $q$:   $i_1 = \arg\min_i \lVert q - k_i \rVert$.  
We then compute the residual as $r_1 = q - k_{i_1}$. For $j = 2, \dots, k$, we iteratively select:  
$i_j = \arg\min_{i \notin \{i_1, \dots, i_{j-1}\}} \lVert r_{j-1} - k_i \rVert$,  
and update the residual as $r_j = r_{j-1} - k_{i_j}$.  
The final prompt consists of the values corresponding to the selected indices $\{i_1, \dots, i_k\}$. Our intuition is that selecting keys based on minimizing the residuals allows us to better represent $q$ using the key pool while also encouraging token diversity during selection.
\\
\\
To ensure effective learning of the key pool, we add an auxiliary loss term to the overall training objective as 
$\mathcal{L} = \mathcal{L}_{\text{next-token}} + \alpha \mathcal{L}_{\text{key}},$
where $\alpha$ is a scaling factor. For the similarity-based strategy, we define the key loss as the sum of distances between the input representation $q$ and the selected keys, i.e., $\mathcal{L}_{\text{key}} = \sum_{i \in \mathcal{K}} \lVert q - k_i \rVert$,
so that the selected keys are encouraged to be close to $q$.  For the attention-based strategy, we use an entropy-based loss defined as  $\mathcal{L}_{\text{key}} = -\sum_{i \in \mathcal{K}} \alpha_i \log \alpha_i$, which promotes a balanced assignment of attention weights. For the residual-based strategy, the key loss is computed as $\mathcal{L}_{\text{key}} = \sum_{j=1}^{k} \lVert r_j \rVert$, which sums the residuals from each iteration. This formulation ensures that the keys selected in successive iterations collectively approximate $q$ as closely as possible.
\\
\\
\noindent \textbf{Stochastic Selection:} Inspired by the stochastic soft prompt approach introduced in \cite{gupta2025stopmodelingasynchronoustime}, this method samples a different prompt size $k$ in every training batch as $k \sim \mathcal{U}(1, P)$, where $\mathcal{U}$ denotes the uniform distribution over $\{1, 2, \dots, P\}$. \cite{gupta2025stopmodelingasynchronoustime} show  that this training paradigm often improves performance and results in learned tokens forming a coarse-to-fine representation. Additionally, it allows for flexible prompt lengths at inference, making it adaptable to varying resource constraints. We also train all the above selection strategies with their corresponding stochastic versions.

\section{Experiments}
The instruction tuning data used for training include multiple tasks, including Automatic Speech Recognition (ASR), automatic speech translation from English to Chinese (En2Zh), emotion recognition (ER), speaker verification (SV), speech question answering (SQA), and audio captioning (ACAP). The datasets used in our experiments, along with their evaluation metrics, are listed in Table \ref{tab:dataset}.

 For a fair comparison, all methods are trained for the same number of iterations: we use a training batch size of 8 and run 90K iterations over the training dataset. Training is conducted on a single H100 GPU, with 90K iterations taking approximately 15–20 hours for most methods. For the dynamic prompt selection methods, we use a prompt pool of 400 tokens and set the prompt size to 160 tokens. For stochastic experiments, we also use 160 tokens at inference for a fair comparison, except for the configuration in row 6 of Table \ref{tab:strategy_comparison}, where 10 tokens are used at inference. We optimize using AdamW and schedule the learning rate with a warmup cosine scheduler, with a warmup period of 3000 steps and a maximum learning rate of $3 \times 10^{-5}$. We employ the LLaMA 8B instruct model\footnote{\url{https://huggingface.co/meta-llama/Meta-Llama-3-8B-Instruct}} as the LLM backbone. We use the same settings as SALMONN for configuring LoRA, Q-Former, and BEATs.

\begin{table}[!t]
    \centering
    \caption{Training data and evaluation metrics used in our multi-task learning experiments. We use LibriSpeech \textit{test-clean} as the ASR test set and IEMOCAP Session 5 as the test set for ER. For the remaining datasets, we follow the test splits officially provided.}

    \scalebox{0.9}{
    \resizebox{\linewidth}{!}{%
    \begin{tabular}{lcc}
        \specialrule{.15em}{.2em}{.1em}
         Task & Dataset & Metric(s) \\
        \midrule
          ASR & Librispeech \textit{train-clean-100} \cite{7178964} & Word Error Rate (WER) \\
          En2Zh & CommonVoice (CV) \cite{ardila-etal-2020-common} & Character  Error Rate (CER) \\
          ER & IEMOCAP \cite{busso2008iemocap} & Accuracy (ACC) \\
          SV & VoxCeleb1 \cite{nagrani2017voxceleb} &Accuracy (ACC)\\
          SQA & LibriSQA-Part $\mathrm{I}$ \cite{zhao2023librisqa} & Rouge-L , BLEU \\
          ACAP & Clotho \cite{drossos2020clotho} & Rouge-L , BLEU \\
        \specialrule{.15em}{.2em}{.1em}
    \end{tabular}%
    }
    }
    \label{tab:dataset}
\end{table}

\begin{figure}[!t]
    \centering
        \scalebox{0.9}{
    \includegraphics[width=\linewidth]{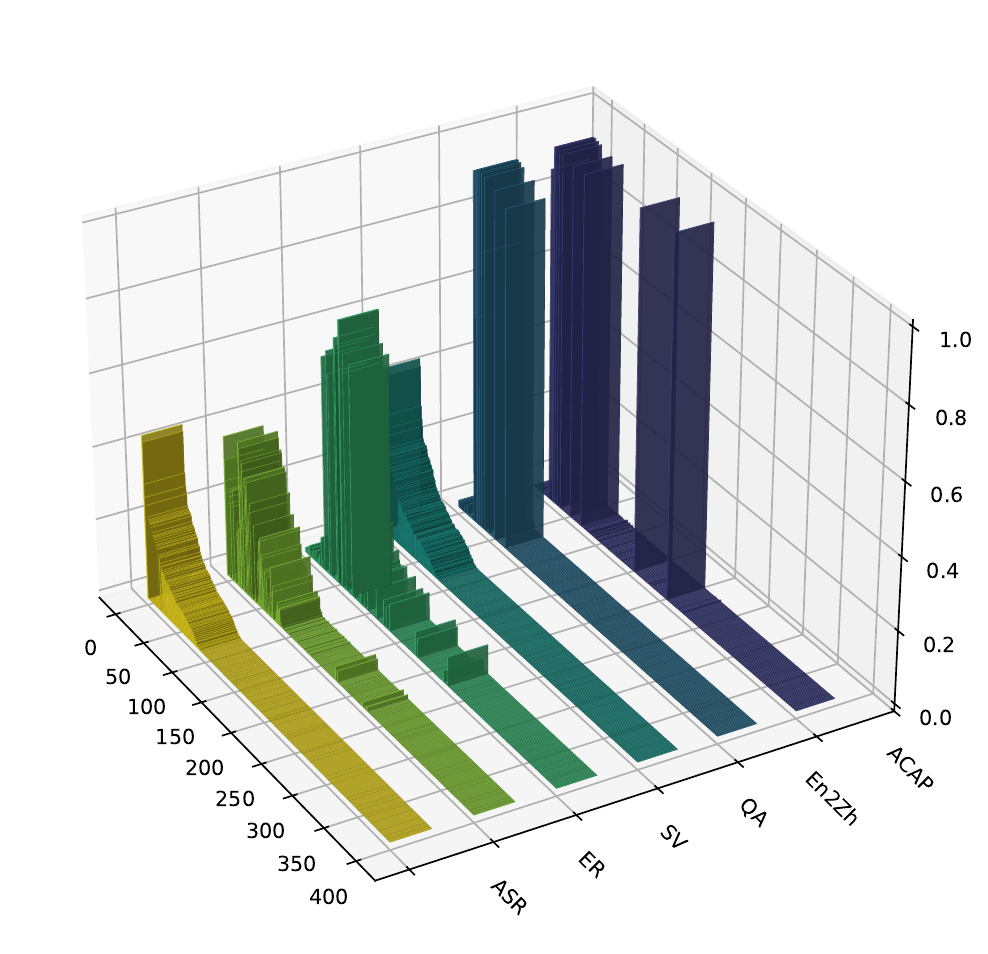}
    }
    \caption{Token usage distribution across tasks in the test set. The z-axis represents token frequency, the x-axis shows token indices sorted by frequency for ASR, and the y-axis indicates the task. A prompt pool of 400 tokens was used, with each instance selecting 10 tokens at inference.}
    \label{fig:token_task_diversity}
\end{figure}

\section{Results and Analysis}
 In \autoref{tab:strategy_comparison}, we compare LoRA, soft prompting (or prefix tuning), and our proposed dynamic prompt selection strategies for adapting a backbone LLM for audio-text multitask learning. We include SALMONN's performance on our dataset for comparison, with results aligning with its benchmarking in \cite{wang2024audiobench}. However, a direct comparison with SALMONN is not entirely valid, as our model is trained on fewer tasks with a smaller dataset, while SALMONN uses a larger 13B LLM backbone\footnote{\url{https://huggingface.co/lmsys/vicuna-13b-v1.5}}.

\begin{table*}[!t]
    \centering
    \caption{Performance comparison of different strategies across tasks (\textbf{best} and \underline{second-best} values computed ignoring the SALMONN row). We compare LoRA, soft prompting, and Dynamic Prompt Selection (DPS). Results are shown for various prompt selection strategies from Section \ref{prompt_selection_stratgies}, trained with and without stochasticity. A prompt pool of 400 tokens is used, with 160 tokens at inference, except in row 6 ($^*$), where only 10 tokens are used.}

    \scalebox{0.85}{
    \resizebox{\textwidth}{!}{%
    \begin{tabular}{lcccccccc}
        \specialrule{.15em}{.2em}{.1em}
         Method & ER & \multicolumn{2}{c}{QA} & ASR & SV & En2Zh & \multicolumn{2}{c}{AAC} \\
        \midrule
                 & ACC $\uparrow$ & RougeL $\uparrow$  & BLEU $\uparrow$  & WER $\downarrow$  & ACC $\uparrow$  & CER $\downarrow$ & RougeL $\uparrow$  & BLEU  $\uparrow$ \\
        \midrule
        LORA & 0.5542 & 0.4462 & 0.2562 & 0.0840 & 0.8915 & 0.8024 & 0.3587 & 0.1320 \\
        Soft prompt & 0.5848 & \underline{0.4997} & 0.2838 & 0.0447 & 0.8860 & 0.6848 & 0.3557 & 0.1310 \\
        Stochastic Soft Prompt  & 0.6065 & 0.4996 & 0.2844 & \textbf{0.0421} & 0.8860 & \textbf{0.6639} & \underline{0.3605} & 0.1320 \\
        DPS, similarity & \underline{0.6462} & 0.4924 & 0.2808 & \underline{0.0426} & 0.8942 & \underline{0.6658} & 0.3578 & 0.1288 \\
        DPS, similarity, stochastic & 0.6119 & \textbf{0.4998} & 0.2838 & \textbf{0.0421} & 0.8936 & 0.6736 & \textbf{0.3613} & \textbf{0.1344} \\
        DPS, similarity, stochastic, 10$^*$ & 0.5975 & 0.4991 & 0.2834 & 0.0491 & 0.8910 & 0.6844 & 0.3594 & 0.1292 \\
        DPS, attention & 0.4675 & 0.4915 & \textbf{0.2874} & 0.0673 & 0.8916 & 0.6882 & 0.3564 & 0.1309 \\
        DPS, attention, stochastic & 0.4350 & 0.4438 & 0.2312 & 0.1411 & \textbf{0.8951} & 0.9004 & 0.3415 & 0.1182 \\
        DPS, residual & 0.6119 & 0.4929 & \underline{0.2859} & 0.0832 & \underline{0.8950} & 0.7066 & 0.3444 & 0.1158 \\
        DPS, residual, stochastic & \textbf{0.6480} & 0.4818 & 0.2778 & \underline{0.0511} & 0.8945 & 0.6904 & 0.3580 & \underline{0.1326} \\
        \specialrule{.15em}{.2em}{.1em}
        SALMONN & 0.7401 & 0.3658 & 0.1184 & 0.6143 & 0.9365 & 0.7798 & 0.3299 & 0.1080 \\
        \specialrule{.15em}{.2em}{.1em}
    \end{tabular}%
    }
    }
    \label{tab:strategy_comparison}
\end{table*}

\subsection{Comparison of Different Fine-tuning Strategies}

Among the dynamic prompt selection strategies we experiment with, we observe that similarity-based selection performs the best, achieving either the best or second-best results on 6 out of the 8 tracked metrics. This is followed by the residual prompt selection strategy. Our experiments are conducted with a prompt pool of size 400, leading to approximately 3.3M trainable parameters for dynamic prompt selection, which is comparable to LoRA for a fair comparison. However, we observe in our experiments that the prompt pool can be significantly reduced without substantial performance loss, allowing for much fewer trainable parameters. Overall, prompt-based methods consistently outperform LoRA while providing greater flexibility and potentially requiring far fewer trainable parameters. 

While non-stochastic versions generally perform better than their stochastic counterparts on average, the stochastic version offers the flexibility to use a much smaller prompt size at inference. This is demonstrated in the row marked with *, where inference is done with only 10 tokens instead of 160 while still achieving comparable performance.


Overall, our results show that dynamic prompt selection outperforms LoRA for multitask fine-tuning, requires significantly less data, and provides a more interpretable fine-tuning method. It also offers greater controllability compared to LoRA and soft prompts, making it a promising alternative for data efficient single stage adaptation of LLMs to diverse speech and audio tasks.
\subsection{Prompt Diversity Analysis}
Using prompt pools allows each instance to select its own tokens for the task, rather than relying on a fixed set of tokens like in standard soft prompts. This makes the approach more adaptable, as different instances can use different tokens based on what they need. Additionally, training this approach in a stochastic manner helps keep the prompt size small while still ensuring that the selected tokens are effective. \autoref{fig:token_task_diversity} shows the distribution of token usage across tasks when each instance selects 10 tokens during inference. We observe:
\begin{itemize}
    \item \textbf{Token Diversity}: Although each instance picks only 10 tokens, the total number of distinct tokens used across the test set is much higher. This suggests that instances are making different choices, leading to more varied token usage. In contrast, other PEFT methods use the same set of tokens across all instances.
    \item \textbf{Token Diversity Across Tasks}: Similar tasks tend to rely on overlapping tokens, while tasks that are different from each other use more distinct sets. For example, ASR and QA share a large portion of their tokens, whereas Audio Captioning selects a very different set. On the other hand, Emotion Recognition uses some tokens that overlap with ASR, which makes sense since both tasks involve processing speech. In contrast, Audio Captioning deals with non-speech sounds, so it requires a different selection of tokens.
    \item \textbf{Explainability} The prompt pool makes this prompting method more interpretable compared to other PEFT approaches. By analyzing the token distribution used by a new task, we can better understand how it relates to existing tasks. This could also help in generalizing to unseen tasks, as a new task instance can select already trained tokens that are most relevant to its needs, rather than relying on a fixed prompt or fixed LoRA adapters.
\end{itemize}
\subsection{Prompt Size Analysis}

A key advantage of the stochastic version of dynamic prompt selection methods is that at inference, we can select a smaller prompt length without significantly compromising performance. This is demonstrated in Table \ref{tab:strategy_comparison}, comparing rows 5 and 6. Row 5 corresponds to inference with a prompt length of 160, while row 6 represents inference with a significantly reduced prompt length of 10. While performance decreases across all tasks, the drop is relatively small. This suggests that stochastic training enables more efficient inference, reducing computational costs while maintaining strong performance. Notably, even with a prompt length of 10, the model outperforms LoRA on almost all tracked metrics.

\section{Conclusion}
In this work, we explored dynamic prompt selection as an efficient alternative to LoRA and soft prompting for adapting LLMs to audio-language tasks. Our results show that similarity-based dynamic prompt selection, trained in a stochastic setting, achieves the best performance across multiple tasks while enabling significantly lower inference costs without substantial performance degradation. Our method offers greater flexibility and enhanced interpretability. Future work includes exploring its scalability to new tasks and further optimizing prompt selection strategies.

\bibliographystyle{IEEEtran}
\bibliography{mybib}

\end{document}